\documentclass[runningheads]{llncs}
\usepackage[T1]{fontenc}
\usepackage{graphicx}
\usepackage{subcaption}
\usepackage{algorithm}
\usepackage{algpseudocode}
\usepackage{diagbox}
\usepackage{mathtools}                        
\usepackage{siunitx}
\usepackage[table,xcdraw]{xcolor}
\usepackage{multirow}

\usepackage[colorlinks=true, allcolors=blue]{hyperref}

\begin{document}
\title{Mean Opinion Score as a New Metric for User-Evaluation of XAI Methods\thanks{Supported by organization Laboratoire Bordelais de Recherche en Informatique.}}
\author{Hyeon Yu \and
Jenny Benois-Pineau\inst{1}\orcidID{0000-0003-0659-8894}
\and
Romain Bourqui\inst{1}\orcidID{0000-0002-1847-2589} 
\and 
Romain Giot\inst{1}\orcidID{0000-0002-0638-7504}
\and 
Alexey Zhukov\inst{1}\orcidID{0009-0009-0518-5474}
}
\authorrunning{H. Yu et al.}

\institute{University of Bordeaux, Bordeaux, Nouvelle-Aquitaine, France
\email{\{first\}.\{last\}@u-bordeaux.fr}}


\maketitle              % typeset the header of the contribution

% \chapter*{Abstract}
\begin{abstract}

% should be 10 lines

This paper investigates the use of Mean Opinion Score (MOS), a common image quality metric, as a user-centric evaluation metric for XAI post-hoc explainers. To measure the MOS, a user experiment is proposed, which has been conducted with explanation maps of intentionally distorted images. Three methods from the family of feature attribution methods – Gradient-weighted Class Activation Mapping (Grad-CAM), Multi-Layered Feature Explanation Method (MLFEM), and Feature Explanation Method (FEM) – are compared with this metric. 
Additionally, the correlation of this new user-centric metric with automatic metrics is studied via Spearman's rank correlation coefficient. MOS of MLFEM shows the highest correlation with automatic metrics of Insertion Area Under Curve (IAUC) and Deletion Area Under Curve (DAUC). However, the overall correlations are limited, which highlights the lack of consensus between automatic and user-centric metrics.


\end{abstract}

% \clearpage

\section{Introduction}
Artificial intelligence (AI) systems are rapidly becoming ubiquitous, influencing decisions in diverse domains ranging from medicine to finance.  These models function as ``black-box'' due to their inherent complexity, making it difficult to explain their reasoning and potentially perpetuating biases in the decision-making process. This growing reliance, however, has exposed a critical gap in transparency and interpretability.

Regulations such as the General Data Protection Regulation (GDPR) \cite{EU-GDPR-2016}, and the need to address algorithmic bias and fairness highlight the importance of understanding how these complex systems, often deep learning models, arrive at their conclusions \cite{Goodman_2017}. The Defense Advanced Research Projects Agency (DARPA) \cite{Gunning_Aha_2019} recognizes the critical role of eXplainable AI (XAI) in the success of deep learning. Users may reject models they don't comprehend, leading to decreased system adoption and potential ethical concerns \cite{https://doi.org/10.1002/bdm.2155}, \cite{Dietvorst2015}. 

% While various explanation methods for image classification have emerged, just to mention a few \cite{DBLP:journals/corr/SelvarajuDVCPB16}, \cite{bourroux:hal-03689004}, \cite{DBLP:journals/corr/RibeiroSG16}, and \cite{DBLP:journals/corr/LundbergL17},
While various explanation methods for image classification have emerged (e.g. \cite{DBLP:journals/corr/SelvarajuDVCPB16}, \cite{bourroux:hal-03689004}, \cite{DBLP:journals/corr/RibeiroSG16}), their effectiveness in fostering human understanding remains uncertain. The contributions of this work are the following: 
\begin{itemize}
    \item We propose the use of Mean Opinion Score (MOS), which has been widely used in image quality assessment \cite{ITUR}, as a new metric for post-hoc XAI evaluation involving user experience;
    \item We design the protocol of its collection for a ``hard'' case of distorted images and well- and poorly classified situations with a trained model; 
    \item We apply this metric to three explainers from the feature attribution family, one of which being popular Grad-CAM \cite{DBLP:journals/corr/SelvarajuDVCPB16} and two are statistical explainers proposed in previous works;
    \item Finally, we study the correlation of this metric with automatic metrics ``Insertion Area Under Curve'' (IAUC) and ``Deletion Area Under Curve'' (DAUC) \cite{petsiuk2018rise}. 
\end{itemize}

%In our work, we propose a new evaluation metric for assessment of XAI methods, Mean Opinion Score (MOS), which has been widely used in visual quality assessment \cite{ITUR}. The rationale behind using it for XAI evaluation is that while remaining a metric with user involvement, it requires a lower cognitive load and time spent to evaluate XAI. Amongst all XAI methods available up to now, we focus on three specific post-hoc methods: FEM \cite{ahmedasiffuad:hal-02963298}, Multilayered FEM (MLFEM) \cite{bourroux:hal-03689004}, and Grad-CAM \cite{DBLP:journals/corr/SelvarajuDVCPB16}. Applied for explanation of classifiers of images, they produce explanation maps with scores $s(x,y)\in R^{[0;1]}$ of pixels. Such a map can then be visualized as a heat map \cite{Zhukov_Benois-Pineau_Giot_2023}. 

%The two first methods have been developed in our previous work, Grad-CAM remains the most popular method, namely to explain decision-making on medical images \cite{ALLGAIER2023102616}.  We also study the correlation of MOS with widely spread "automatic metrics", "Insertion Area Under Curve" (IAUC) and "Deletion Area Under Curve" (DAUC). 
The remainder of the paper is organized as follows. In section \ref{sec:Related_Work} related work is reviewed. In section \ref{sec:PVE} the design of the assessment experiment is proposed. Section \ref{sec:SA} is devoted to statistical analysis of experimental data, while in section \ref{sec:result-SA} we compare both the methods in terms of the MOS metric and present the results of the study of the correlation of MOS with IAUC and DAUC. Section \ref{sec:conclusion} concludes our work and outlines its perspectives. 

% \clearpage % You need \clearpage at the end of every chapter to force images included in this chapter to be rendered somewhere within it
\section{Related Work}
\label{sec:Related_Work}

%%%%%%%%%%%%%%%%%%%%%%%%%%%%%%%%%%%%%%%%%%%%%%%%%%%%%%%%%%%%%%%%%%%%%%%%
% \subsubsection{Evaluation Metrics for Explanation Methods with User Involvement} \label{sec:XAI-metrics}
Nowadays, there is no consensus on the evaluation metrics for explanation methods. We propose to devise them into i)automatic or no-reference metrics, ii) semi-user-centered or reference-based metrics, and iii) fully user-centered metrics. 
\subsubsection{No-Reference Metrics.} 
No-reference metrics do not depend on human judgments \cite{Zhukov_Benois-Pineau_Giot_2023}. Examples of these metrics are  Insertion Area Under the Curve (IAUC) and Deletion Area Under Curve (DAUC) \cite{petsiuk2018rise}. DAUC evaluates explanation maps by progressively masking image regions based on their importance supplied by the explanation method (highest score pixels masked first). It calculates the area under the curve of the image's classification score as the masked proportion increases. A lower DAUC score indicates a better explanation because masking supposedly important regions should significantly reduce the initial score. IAUC works similarly, but in the opposite direction: it measures how revealing important areas increases the score.
\subsubsection{Reference-Based Metrics.} 
Reference-based metrics serve for evaluation of a XAI method by human interpretations or comparisons with human interest expressed a priori by Gaze Fixation Density Maps (GFDMs) \cite{bourroux:hal-03689004}. Such metrics are Pearson correlation coefficient (PCC) and similarity (SIM) measured between two maps. In the paper \cite{Zhukov_Benois-Pineau_Giot_2023}, the authors evaluate the explanation methods Grad-CAM \cite{DBLP:journals/corr/SelvarajuDVCPB16}, MLFEM \cite{bourroux:hal-03689004}, and FEM \cite{ahmedasiffuad:hal-02963298} using both reference-based and no-reference metrics. 
We qualify ``reference-based metrics'' from \cite{Zhukov_Benois-Pineau_Giot_2023} as semi-automatic, as they do not require the assessment of explanation maps by the user himself, but a comparison of the explanation maps with the importance of pixels in the input image for the user.
\subsubsection{Fully User-Centered Metrics.} 
On the contrary, \cite{KENNY2021103459} focuses on assessing explanation methods through user perceptions of the accuracy of AI classifications. This approach is therefore fully user-centered. The authors conduct user studies focused on both item- and system-level metrics to evaluate the performance and acceptance of AI classifications. Item-level measures focus on individual classifications. Participants judge the correctness and reasonableness of each classification on a 5-point scale, using their understanding of these terms. These ratings provide a detailed view of how participants perceive the accuracy and logic behind each classification. System-level measures evaluate users' overall trust and satisfaction with the entire AI system. These metrics align with approaches used in the DARPA XAI program \cite{Gunning_Aha_2019}, where user experience with explanations is assessed to gauge trust and satisfaction in the system itself. Standardized questionnaires are employed for this purpose. 

\subsubsection{Mean Opinion Score.}

The Recommendation of International Telecommunication Union Telecommunication Standardization Sector (ITU-T) P.800.2 \cite{ITU-P} provides a detailed discussion on the differential perception of audio and video quality by humans, highlighting the absence of a singular definition for a 'good' quality. To evaluate user experience with new audio or video systems, subjective experiments are conducted in which participants rate the performance of these systems. These experiments employ rating scales to convert user opinions into numerical scores, \textit{opinion scores (OS)},  which can then be averaged to produce a Mean Opinion Score (MOS) per content item. The most widely utilized scale is the 5-point Absolute Category Rating (ACR) scale, where participants independently rate each sample without any reference. Alternatively, participants may be asked to compare a processed sample to an original one. Furthermore, the 5-point ACR scale serves as the primary quality metric in the single- and double-stimulus subjective assessment of television picture quality \cite{ITUR}. ACR is typically the same scale as the famous Likert scale in its 5-point version \cite{Likert_1932}, which is employed for user studies. Hence, in the following design of our experiment for MOS computation, we will employ the Likert scale. 
% ITU-T P.800.2 

% \section{Psycho Visual Experiment} \label{PVE}

\section{MOS as a Quality Metric for XAI Methods} \label{sec:PVE}

Drawing inspiration from the established Mean Opinion Score (MOS) metric used in image quality assessment \cite{ITUR}, we explore the potential of applying MOS to evaluate the quality of explanation methods in XAI. 

\subsubsection{Mean Opinion Score (MOS).}
MOS provides a standardized way to quantify perceived quality through subjective human judgments. By adapting this approach to the realm of explanation methods, we aim to develop a reliable metric that involves user feedback to assess the effectiveness and clarity of explanations generated by different methods. 

MOS in the realm of explanation methods, refers to the mean rating participants provided for the explanation map given by the explanation method for each image. For all participants $j$, $j \in \{1,2,3,...,N\}$, MOS of image $I_i$ and explanation method $k$ can be expressed as:
\begin{equation} 
    MOS_{{I_i}{k}} = \frac{1}{N} \sum_{j=1}^{N} o_{ijk}
\end{equation}
where $o_{ijk}$ is the opinion score of participant $j$ for image $I_i$ and explanation method $k$.

% of the image $i$, of participant $j$ for the explanation method $k$.
The opinion score $o$ ranges from 1 to 5 with 5 corresponding to ``excellent''. 

To obtain the MOS of the explanation maps of explanation methods of images, we have to design a user-centric experiment, which we present in the following. 

\subsubsection{Design of Psycho-Visual Experiment to Measure MOS of Explanation Maps of XAI Methods.} \label{sec:Methods-PVE}
This user study departs from traditional approaches that focus on usefulness and satisfaction for users \cite{ford2022explaining}. Instead, it adopts a methodology akin to visual quality assessment, commonly employed for evaluating images and videos \cite{ITUR}. The goal of the psycho-visual experiment is to evaluate the quality of explanation maps generated by different explanation methods. Participants are thus tasked with rating the explanation maps associated with distorted images, all of which were classified by the same Deep Neural Network (DNN) model. We have taken the ResNet50 classifier as in \cite{Zhukov_Benois-Pineau_Giot_2023}. The experiment consisted of sessions, conducted either online or offline. Participants chose their preferred format at the outset and remained in that format for the entire experiment. For online sessions, screen sharing was utilized for participant supervision. The offline sessions were conducted under supervision in the presence of the coordinator.

%This experiment draws inspiration from the work of Zhukov et al. \cite{Zhukov_Benois-Pineau_Giot_2023}, where a classification model identical to ours was used alongside the same explanation methods and a subset of their experiment images. 
\subsubsection{Data.}
\begin{figure}[ht!]
    \centering
     \begin{subfigure}[b]{\textwidth}
        \begin{subfigure}[b]{0.3\textwidth}
            \includegraphics[width=\linewidth]{images/methods/Degraded/gau-noise/1-k-25-0.jpg}  
        \end{subfigure}
        \hfill
        \begin{subfigure}[b]{0.3\textwidth}
            \includegraphics[width=\linewidth]{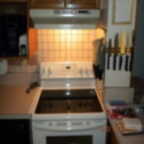}  
        \end{subfigure}
        \hfill
        \begin{subfigure}[b]{0.3\textwidth}
            \includegraphics[width=\linewidth]{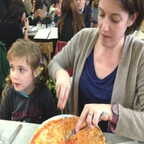}  
       \end{subfigure}
       \caption{Weak distortion}
    \end{subfigure}
    \begin{subfigure}[b]{\textwidth}
        \begin{subfigure}[b]{0.3\textwidth}
            \includegraphics[width=\linewidth]{images/methods/Degraded/gau-noise/1-k-100-0.jpg} 
        \end{subfigure}
        \hfill
        \begin{subfigure}[b]{0.3\textwidth}
            \includegraphics[width=\linewidth]{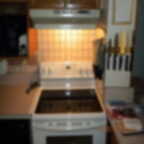} 
        \end{subfigure}
        \hfill
        \begin{subfigure}[b]{0.3\textwidth}
            \includegraphics[width=\linewidth]{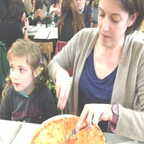} 
       \end{subfigure}
       \caption{Strong distortion}
    \end{subfigure}
    \caption{Weakly and strongly distorted images of SALICON dataset: additive Gaussian noise (left), Gaussian blur (center), and uniform random brightness shift (right)}
    \vspace{-0.5cm}
    \label{fig:disorted-images}
\end{figure}

From the whole set of distorted images from \cite{Zhukov_Benois-Pineau_Giot_2023}, which we used only two levels of distortions, we qualify as ``weak'' and ``strong'' (see example in Fig.~\ref{fig:disorted-images}). The types of distortions were: additive Gaussian noise, Gaussian blur, and uniform random brightness shift.  %our experiment utilized images distorted with various manipulations shown in : additive Gaussian noise (k = 25, 100), Gaussian blur (k = 1.25, 2), and uniform brightness distortion (k = 25, 200).%
This subset selection aimed to manage the experiment's duration. In addition, only distorted images were used since we are specifically interested in the performance of the explanation methods on perturbed data. These images were pushed through the trained model and explanation maps obtained by XAI methods were submitted to user evaluation. 
In total, 300 distorted images were used, and three different XAI methods were applied to each image to obtain the explanation maps, resulting in a total of 900 maps to evaluate.

%As the image classifier ResNet50 \cite{he2015deep} was used as  in \cite{Zhukov_Benois-Pineau_Giot_2023}. More information on the classification model and creation of the distorted images can be found in . 

%In addition, the software supporting the experiment was developed by A. Zhukov during his internship at LaBRI and then adjusted by ourselves to adapt it for both modes of the experiment: the offline and online modes. 

\subsubsection{Explanation Methods.}
Among explainable AI methods for image classification, Gradient-weighted Class Activation Mapping (Grad-CAM) \cite{DBLP:journals/corr/SelvarajuDVCPB16} is the most widely used approach. Recently, Feature-wise Explanation Maps (FEM) and Multi-layer Feature-wise Explanation Maps (MLFEM) were proposed as alternative techniques, demonstrating superior performance to Grad-CAM \cite{bourroux:hal-03689004}. These methods fall under the category of Feature Attribution methods. Specifically, they assign scores to individual features or pixels within the input image to show their contribution into the classification decision. These scores represent the relative importance of each feature in influencing the model's final prediction. While Grad-CAM relies on gradients to evaluate feature significance, FEM employs a k-sigma rule on the activation values of the last convolutional layer. MLFEM, on the other hand, extends this analysis to all convolutional layers within the model.

\subsubsection{Experimental Protocol.} \label{ssec:exProtocol}
\begin{figure}[ht!]
        \includegraphics[width=\linewidth]{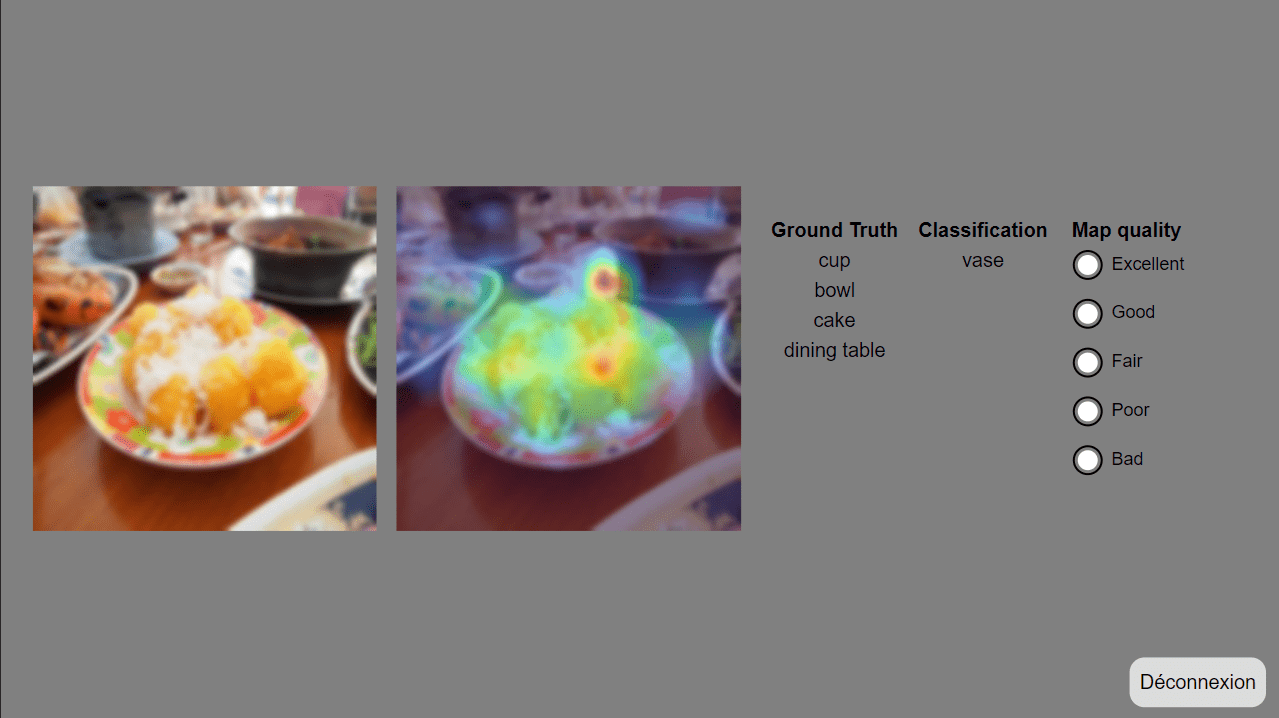}
        \caption{Evaluation screen. From left to right: distorted image, importance heat map, ground truth class labels, model prediction result, and Likert scale for explanation quality.}
        \label{fig:experiment-screen}   
\end{figure}

The ITU-R BT.500-15 recommendation \cite{ITUR} for single-stimulus image quality assessment was followed. Indeed, the users had to evaluate the quality of only one image - the explanation map overlaid on the original image as a heat map.  The recommendation ITU-R BT.500-15 considers visual fatigue and memory limitations, advising a 10-second exposure time per image. To allow the participants to take in the information on the screen and the time to click on the map quality, the exposure time was extended to 20 seconds per image pair.
Between each map screen, a 5-second inter-stimulus interval (ISI) with a mid-gray screen was used. ISI minimizes carry-over effects, allowing participants to reset their visual attention and reduce fatigue. Each session presented participants with 225 images.
The duration of sessions was designed to avoid visual fatigue of the subjects. Thus, the time spent for one session per subject was approximately 1 hour 20 minutes if the subjects worked without interruption. There was also an option to interrupt the session and restart it with unseen images. The overall participation time per subject was about 5 hours and 20 min. The experiment screen is depicted in Fig.~\ref{fig:experiment-screen}. 

\subsubsection{Evaluation Rules.}
As illustrated in Fig.~\ref{fig:experiment-screen}, adapting the commonly used ITU-R scale ~\cite{ITUR}, the experiment employed a 5-point Likert scale with options: Excellent, Good, Fair, Poor, and Bad.

An explanation is considered \textit{good} if the explanation map and the classification are consistent: the classification corresponds to the image's content highlighted by the XAI method. %see illustration in (Fig.~\ref{fig:def-likert-a}). 
Another case of a good map is when the classification is incorrect (poorly classified) and the heat map highlights pixels outside the ground truth objects. %(Fig.~\ref{fig:def-likert-b}).

An explanation is considered \textit{bad} if the explanation map and the classification are inconsistent: the classification is correct (well-classified), but the relevant pixels are not highlighted %(Fig.~\ref{fig:def-likert-c}) 
or the classification is incorrect (poorly classified) but the pixels corresponding to the true class are highlighted.

\subsubsection{Participants Recruitment.}
% Participants have to have an acute visual attention and a quick processing speed to evaluate explanation maps. These abilities decline with age \cite{doi:10.1111/j.1467-8721.2007.00478.x}, \cite{Ebaid2019}, \cite{Groth2000}, \cite{Lachica2023.05.18.540524}, \cite{doi:10.2352/EI.2023.35.8.IQSP-302}, \cite{Salthouse1996}. 
Participants have to have acute visual attention and a quick processing speed to evaluate explanation maps. These abilities decline with age \cite{Ebaid2019}, \cite{doi:10.2352/EI.2023.35.8.IQSP-302}. To align with the young adult group from the mentioned papers, participants aged 18 to 29 were recruited for the experiment.
% User studies often categorize participants into young and older adult groups. For our study, participants aged 18 to 29 were recruited, aligning with the young adult group in \cite{Ebaid2019} and \cite{doi:10.2352/EI.2023.35.8.IQSP-302}.
To ensure participants could accurately assess the heat map of the explanation maps, which rely on color differentiation, all underwent an Ishihara test \cite{ishihara1917tests} for red-green color deficiencies. Only those who passed proceeded to the image evaluation task.

The experiment involved 31 participants. Fifteen participated in person (average age: 25.14 $\pm$ 2.16 y.o.). Sixteen participated online (average age: 25.31 $\pm$ 1.53 y.o.). The online participants were in various locations: France (2), Netherlands (2), Hungary (1), Spain (2), Canada (7), and India (2).

\section{Statistical Analysis of Experiment Data}\label{sec:SA}
This section presents the statistical analysis of the user experiment, with the goals outlined as follows: 
\begin{itemize}
    \item To filter out the most discrepant subjects from the experimental data;
    \item To test the homogeneity of the two groups: online and offline;
    \item To evaluate a set of XAI methods with the proposed MOS metric;
    \item To study the correlation of our fully user-centered metric with automatic metrics IAUC and DAUC. 
\end{itemize}
The results of the statistical analysis are presented in Section~\ref{sec:result-SA}. 

%\subsection{Statistical Analysis}\label{ssec:SA}
%The primary metric for evaluation of XAI methods is the Mean Opinion Score (MOS), which encapsulates the average ratings provided by participants for different explanations of images. 
%To ensure the robustness and reliability of our findings, outlier detection and filtering have been implemented, followed by an examination of the homogeneity of MOS distributions across participant groups using Pearson's chi-squared test of homogeneity \cite{McHugh2013TheChiSquare}. Finally, the analysis involves the application of the Kruskal-Wallis H test \cite{doi:10.1080/01621459.1952.10483441} and the Mann-Whitney U test \cite{10.1214/aoms/1177730491} to discern statistically significant differences in the effectiveness of the explanation methods accordingly to our MOS metric values, considered as the primary statistics issued from the experiment. 

\subsection{Filtering of Outliers} \label{sec:filtO}
We define outliers as subjects whose average opinion scores for all the explanation methods deviate significantly from the average scores of the entire group for that explanation method. Filtering outliers helps to ensure that the evaluation reflects the opinions of a more representative sample and reduces the influence of potentially biased or inattentive participants. Subjects (participants) are excluded if their average opinion score of all the explanation methods falls outside two standard deviations of the average opinion score of the method for the subjects' group.

The filtering process is conducted separately for each group: i) offline $g^{\circ}_{offline}$ and ii) online $g^{\circ}_{online}$. %Here, we denote the two groups as Group, $G^{\circ} = \{g^{\circ}_{offline},g^{\circ}_{online}\}$, where $g^{\circ}_{offline}$ represents participants in the offline group and $g^{\circ}_{online}$ represents participants in the online group. 
As detailed in Section~\ref{sec:Methods-PVE}, the offline group consists of participants who joined the study in person, and in the online group, subjects participated in online sessions.

For each subject within a group, we iterate through each explanation method and perform the following steps to identify outliers:
\begin{enumerate}
    \item \textbf{Calculate individual's statistics}: For a subject $s^{\circ}_j$ in the group $g^\circ$, the average opinion score for the current explanation methods ($e_k$) over $N_I$ images is,
        \begin{equation}\label{eq:AOS}
            \bar{o}(g^{\circ},s^{\circ}_j,e_k) = \frac{1}{N_I}\sum_{i=1}^{N_I} o_{ijk} 
        \end{equation}
    \item  \textbf{Calculate group statistics}: The mean and the standard deviation of the average opinion scores $\bar{o}(g^{\circ},s^{\circ}_j,e_k)$ for the current explanation method are calculated across all subjects of the group.
    % \item \textbf{Define acceptable range}: The acceptable range for the subject's average opinion score is established as two standard deviations away from the mean. 
    % \item \textbf{Comparison of subject's average opinion score}: The subject's average opinion score for the current method is compared to the acceptable range.
    \item \textbf{Identify outliers}: If the subject's average opinion score falls outside the two standard deviations of the mean of the average opinion scores, the subject is identified as an outlier for that particular method and thus removed from the group.
\end{enumerate}

% Subjects identified as outliers for all the explanation methods are added to a list of subjects to be removed from the final analysis. The Subjects' average opinion score can be seen in Fig.~\ref{fig:OS_inds}. Subject 37 from the offline group and subject 24 from the online group were filtered out as outliers following this rule.

Subjects identified as outliers are removed from further analysis. Groups with removed outliers will be denoted as $g_{offline}$ and $g_{online}$.

%  Box-plot of each subject's average opinion score of each explanation method can be seen in Fig.~\ref{fig:OS_inds}. Subject 37 from the offline group and subject 24 from the online group are considered as outliers.
% \begin{figure} [ht!]
%     \centering
%      \includegraphics[width=\linewidth]{images/SA/filtering/all.png}
%     \caption{Box Plot of MOS of Individuals in Each Group. Outliers are identified by the red circles. First row shows: offline group, Second row: online group. From left-to-right: Grad-CAM, MLFEM, FEM.}
%     \label{fig:OS_inds}
    
% \end{figure}
% Following the filtering of outliers, Pearson's chi-squared test of homogeneity \cite{McHugh2013TheChiSquare} was performed on the two groups $g_{offline}$ and $g_{online}$ to check for the possibility of uniting the groups.  
 
\subsection{Test of Homogeneity of MOS for Two Groups}
The objective of this test is to evaluate the feasibility of merging two groups of subjects: the ``online'' and ``offline'' participants. Pearson's chi-square test of homogeneity \cite{McHugh2013TheChiSquare} was performed. If the results indicate homogeneity between these groups, we can compute the Mean Opinion Score (MOS) from the combined opinion scores of all participants for a given image. %  However, if the test suggests a lack of homogeneity, it implies that the groups may have systematically different perceptions of the explanation methods, thus requiring separate analyses of the MOS values.

While the test itself can handle unequal group sizes, the accuracy and reliability of the results can be impacted. To mitigate this concern, the number of subjects in each group was set to the minimum number $N$ of participants in either group ($N = min(|g_{offline}|,|g_{online}|)$). $N$ subjects were randomly selected for the group with a larger sample size ($|g| > N$).

Furthermore, as the chi-squared test requires categorical data, the continuous MOS values were quantized by a uniform quantizer. For a given image $I_i$ a uniform quantizer with a quantization step size, $\triangle$ (in our case, 0.0001) for $MOS_{{I_i}{k}}$ can be expressed as: 
\begin{equation}
    MOSQ_{{I_i}{e_k}} = \triangle\lfloor\frac{MOS_{{I_i}{k}}}{\triangle}+\frac{1}{2} \rfloor
\end{equation}
% By applying Pearson's chi-squared test of homogeneity \cite{McHugh2013TheChiSquare}, with a $\chi^2$ value of 1.24 and a p-value of 0.74, we can infer that the observed distribution of MOS scores between the offline and online groups does not significantly differ from what would be expected by chance. Therefore, it is statistically justifiable to merge the data from both groups for further analysis.

\subsection{Analysis of the Performance of Explanation Methods}
To evaluate the performance of Grad-CAM, MLFEM, and FEM methods with the new metric MOS, a non-parametric statistical analysis was employed on the mean opinion scores (MOS) of explanation maps generated by these methods. The Kruskal-Wallis H test \cite{doi:10.1080/01621459.1952.10483441} was utilized to assess the overall existence of statistically significant differences in MOS ratings across the three methods. Subsequently, pairwise comparisons were conducted using the Mann-Whitney U test \cite{10.1214/aoms/1177730491} to pinpoint specific differences in methods.

The post hoc pairwise Mann-Whitney U test was done with unadjusted p-values rather than employing multiple comparison corrections like Bonferroni. Although common, such corrections can be overly conservative in this context. Since the Kruskal-Wallis test already establishes an overall difference between explanation methods, unadjusted p-values in the pairwise tests allow for a more detailed exploration of these variations. This approach helps identify specific explanation method pairs with significant MOS discrepancies. The analysis result can be seen in Section~\ref{sec:result-SA}.

% \clearpage 
\subsection{Study of correlation of MOS with Automatic Metrics}

It is necessary to position our new user-centric metric with regard to purely automatic metrics, such as IAUC and DAUC. Xu-Darme et al. \cite{xudarme2023stabilitycorrectnessplausibilityvisual} have shown that semi-automatic metrics, which compare explanation maps with gaze fixation density maps did not correlate with DAUC. Hence, our study also examines the correlation between our proposed user-centric metric, MOS, and the automatic metrics IAUC and DAUC. For this, Spearman's rank correlation coefficient is computed between MOS and IAUC and DAUC for the same set of images and across the three XAI methods. The results are detailed in Section \ref{correlation}.

\section{Results and Discussion } \label{sec:result-SA}
% The Code of Studies and Exams recommends the following content for this chapter: "Evaluation, critical analysis of the implemented technical solutions, possibilities for further development."
% However, you are free to structure the content of your thesis as you want
This section presents the results of the statistical analysis of MOS conducted to evaluate the performance of the explanation methods (Grad-CAM \cite{DBLP:journals/corr/SelvarajuDVCPB16}, MLFEM \cite{bourroux:hal-03689004}, and FEM \cite{ahmedasiffuad:hal-02963298}) accordingly to the methodology presented in Section \ref{sec:SA}. Then we present and discuss the results of the study of positioning of our new user-involvement metric with regard to automatic metrics IAUC and DAUC. 

% The analysis focused on Mean Opinion Score (MOS) as the primary user experience metric. Using the Kruskal-Wallis H test \cite{doi:10.1080/01621459.1952.10483441} and post-hoc Mann-Whitney U test \cite{10.1214/aoms/1177730491}, we investigated potential differences in user perception of the explanation methods for well-classified images (labels of the original and distorted images are the same) and poorly classified images (labels of the original and distorted images do not match). In addition, a more detailed comparison of the performance of explanation methods was done for the three distortions of the images. 

%\subsection{Methodlogy}
% Explanation methods' performances were analyzed on well-classified images (labels of the original and distorted images are the same) and poorly classified images (labels of the original and distorted images do not match) and this for the three distortions of the images. 

%If the results of Kruskal-Wallis H test suggest differences between the three explanation methods, then Mann-Whitney parwise test  U test is performed  , if the result suggests differences between the MOS of the images of the explanation methods (rounded to the 4th decimal digits), the explanation methods were compared by comparing the mean MOS values across explanation methods. This can be seen in Tab.~\ref{tab:SA-overall}

\subsubsection{Outlier Filtering}
\begin{figure} [ht!]
    \centering
     \includegraphics[width=\linewidth]{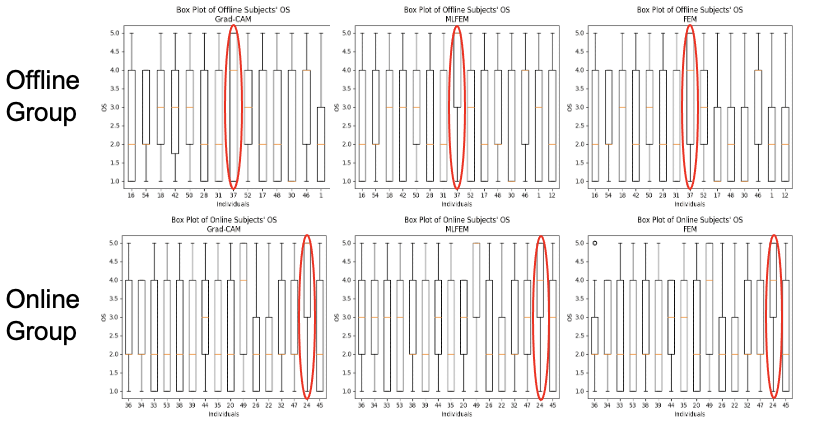}
    \caption{Box Plot of MOS of Individuals in Each Group. The outliers' box-plots have been identified by the red circles. The first row shows the offline group and the second row shows the online group. From left-to-right: Grad-CAM, MLFEM, FEM.}
    \label{fig:OS_inds}
    
\end{figure}

The results of outlier filtering from the participants are illustrated in Fig.~\ref{fig:OS_inds}. It represents the box-plots of each subject's average opinion score of each explanation method Subject 37 from the offline group and subject 24 from the online group have been considered as outliers, denoted by red ellipses.
% Following the filtering of outliers, Pearson's chi-squared test of homogeneity \cite{McHugh2013TheChiSquare} was performed on the two groups $g_{offline}$ and $g_{online}$ to check for the possibility of uniting the groups.  
 
\subsubsection{Homogeneity of Offline and Online Group}
By applying Pearson's chi-squared test of homogeneity \cite{McHugh2013TheChiSquare}, with a $\chi^2$ value of 1.24 and a p-value of 0.74, we can infer that the observed distribution of MOS scores between the offline and online groups with removed outliers ($g_{offline}$ and $g_{online}$) does not significantly differ from what would be expected by chance. Therefore, it is statistically justifiable to merge the data from both groups for further analysis. The following results are presented for the merged groups, totaling 29 participants.

\subsubsection{Tests of Statistically Significance of Differences of All Explanation Methods with MOS Metrics.}
The result of the Kruskal-Wallis H test is shown in Tab.~\ref{tab:Kruskal-result}. We have considered the performances of the explanations on well-classified images and poorly classified images from, each distortion together and separately, independently of distortion level. The bold p-values in the table indicate significant differences in MOS ratings between explanation methods.

\begin{figure}[ht!]
     \centering
     \begin{subfigure}[b]{0.90\linewidth}
         \centering
         \includegraphics[width=\linewidth]{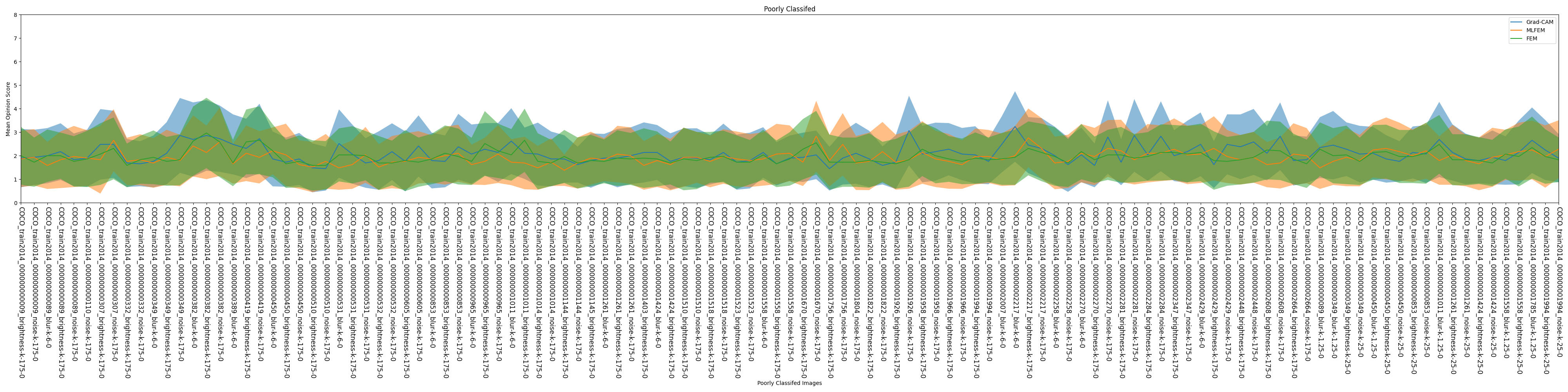}
         \caption{MOS of poorly classified images. Blue: Grad-CAM, green: FEM, orange: MLFEM.} 
         \label{fig:MOSALL-poor}
     \end{subfigure}
     \hfill
     \begin{subfigure}[b]{0.90\linewidth}
         \centering
         \includegraphics[width=\linewidth]{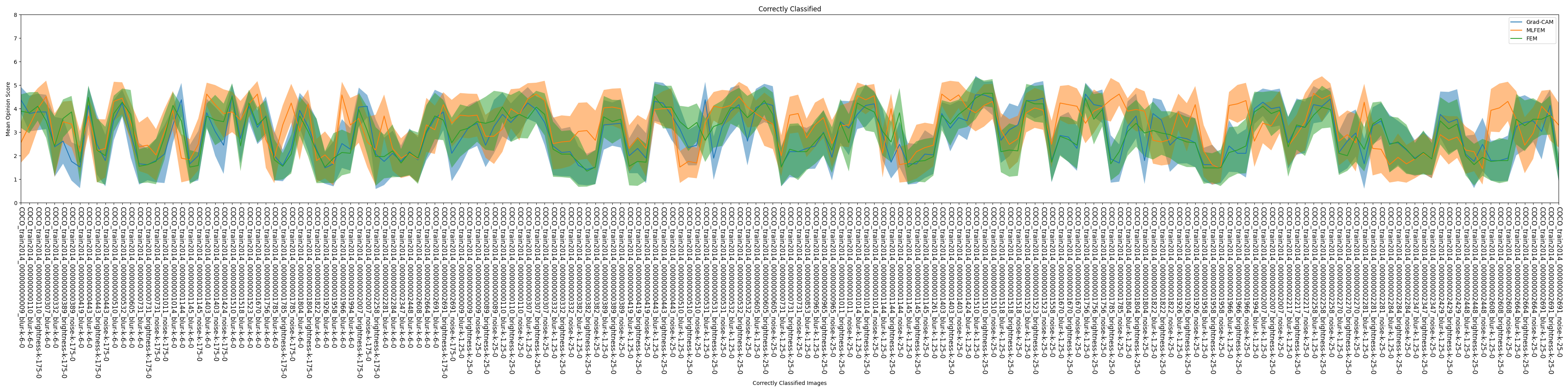}
         \caption{MOS of well-classified images. Blue: Grad-CAM, green: FEM, orange: MLFEM.} 
         \label{fig:MOSALL-well}
     \end{subfigure}
 \caption{MOS and the standard deviation of opinion scores of poorly classified and well-classified images. X-axis: map number, Y-axis: MOS. }
    \label{fig:MOSALL}
    \vspace{-0.5cm}
\end{figure}

This experiment confirms that our newly introduced user-centric metric MOS differentiates between the explanation methods. Note that these methods were already shown to give different results when evaluated with other metrics in \cite{Zhukov_Benois-Pineau_Giot_2023} and \cite{xudarme2023stabilitycorrectnessplausibilityvisual}. 
We illustrate MOS values together with the standard deviation of opinion scores in Fig.~\ref{fig:MOSALL} for all 900 maps we have tested our methods on. Nevertheless, for the distortion such as Gaussian blur on well-classified images and uniform random brightness shift on poorly classified images, there is no significant difference. Therefore, we exclude them from further analysis. It is interesting to note that the MOS values of explanation methods are globally lower for poorly classified images, while the standard deviation of each image is bigger. This can suggest disagreements in rating among the participants.

%For poorly classified images, the average MOS of Grad-CAM is 2.0940 with an average standard deviation of 1.1985. For MLFEM, it is 1.9284 with 1.1510 as the mean standard deviation, and for FEM, shown in green, the average MOS is 1.9611 with a mean standard deviation of 1.1339. For well classified images the average MOS of Grad-CAM is 2.9190 with an average standard deviation of 0.8976. For MLFEM, it is 3.2423 with 0.8889 as the mean standard deviation, and for FEM, shown in green, the average MOS is 2.9360 with a mean standard deviation of 0.8871. Hence our observation is that MOS is generally better for well-classified images. To verify this assumption, we compare the methods with MOS performing Mann-Whitney U test. 

\begin{table}[ht!]
    \centering
    \footnotesize
    \begin{tabular}{|c|c|c|c|c|}        
        \hline
        \rowcolor[HTML]{C0C0C0}
         \multicolumn{1}{|l|}{\backslashbox{Classification}{Distortion}}  & All             & \begin{tabular}[c]{@{}c@{}}Additive Gaussian\\ Noise\end{tabular} & \begin{tabular}[c]{@{}c@{}}Gaussian \\ Blur\end{tabular} & \begin{tabular}[c]{@{}c@{}}Uniform Random \\ Brightness Shift\end{tabular} \\ \hline
Poorly                                          & \textbf{0.0012} & \textbf{0.0125}                                                   & \textbf{0.0004}                                          & 0.4111                                                        \\ \hline
Well                                            & \textbf{0.0006} & \textbf{0.0148}                                                   & 0.2899                                                   & \textbf{0.0412}                                               \\ \hline
\end{tabular}
    \caption{The p-value of the Kruskal-Wallis H test for testing the differences of the XAI methods with MOS metric.  }
    \label{tab:Kruskal-result}
    \vspace{-1.5cm}
\end{table}

\subsubsection{Pair-wise Comparison of XAI Methods with MOS metric.}
The results of the pairwise  Mann-Whitney U test are presented in Tab.~\ref{tab:Mann-Whitney}. The bold p-value indicates that there exists a statistically significant difference between the methods.

Here once more, we distinguish between well-classified and poorly classified images and perform computations for all images, see column three, and for each considered distortion, see columns 4,5,6 of Tab.~\ref{tab:Mann-Whitney}. 

The post-hoc Mann-Whitney U test indicates significant differences in MOS between Grad-CAM and MLFEM (p = 0.0007) and between Grad-CAM and FEM (p = 0.0041) for poorly classified images. Here, the average MOS of Grad-CAM is $2.0940\pm0.3685$. For MLFEM, it is $1.9284\pm0.2635$ and for FEM, $1.9611\pm0.2627$. Therefore, for poorly classified images, Grad-CAM is a better explainer than FEM and MLFEM. 

For well-classified images, the post-hoc test revealed significant differences between MLFEM and both Grad-CAM (p = 0.0008) and FEM (p = 0.0009). MLFEM obtained a statistically higher mean MOS ($3.2423\pm0.9088$) compared to Grad-CAM ($2.9190\pm0.9368$) and FEM ($2.9360\pm0.9171$).
Hence, in the case of well-classified images, MLFEM is the best explainer according to the metric MOS.

More detailed performances of explanation methods on different distortions can be seen in Tab. \ref{tab:SA-overall}. For poorly classified images, Grad-CAM exhibits higher mean MOS, achieving $2.1304\pm0.4061$ for images with additive Gaussian noise and $2.2714\pm0.3689$ for images with Gaussian blur, which further supports the Grad-CAM's superior performance on poorly classified images. In addition, the superiority of MLFEM on well-classified images is demonstrated, as it outperforms Grad-CAM and FEM on well-classified images with additive Gaussian noise ($3.2471\pm0.8902$) and uniform random brightness shift ($3.3508\pm0.8353$).

%%%%%%%%%%%%%%%%%%%%%%%%%%%%%%%%%%%%%%%%%%%%%%%%%%%%%%%%%%%%%%

% For well classified images the average MOS of Grad-CAM is 2.9190 with an average standard deviation of 0.8976. For MLFEM, $3.2423\pm0.9088$, and for FEM, shown in green, the average MOS is 2.9360 with a mean standard deviation of 0.8871.

% Grad-CAM achieved a statistically higher mean MOS ($2.0940\pm0.3685$) compared to MLFEM ($1.9284\pm0.2635$). The mean MOS of FEM was calculated to be $1.9611\pm0.2627$.

\begin{table}[ht!]
    \centering
    \footnotesize
    \begin{tabular}{|c|c|c|c|c|c|}
    \hline
    \rowcolor[HTML]{C0C0C0}
Classification                                                                & \begin{tabular}[c]{@{}c@{}}Explanation \\ Methods\end{tabular}  & All             & \begin{tabular}[c]{@{}c@{}}Additive Gaussian \\ Noise\end{tabular} & \begin{tabular}[c]{@{}c@{}}Gaussian \\ Blur\end{tabular} & \begin{tabular}[c]{@{}c@{}}Uniform Random \\ Brightness Shift\end{tabular} \\ \hline
\multirow{3}{*}{\begin{tabular}[c]{@{}c@{}}Poorly \\ Classified\end{tabular}} & \begin{tabular}[c]{@{}c@{}}Grad-CAM/ \\ MLFEM\end{tabular} & \textbf{0.0007} & 0.1677                                                             & \textbf{0.0003}                                          & N/A                                                           \\ \cline{2-6} 
                                                                              & \begin{tabular}[c]{@{}c@{}}Grad-CAM/ \\ FEM\end{tabular}   & \textbf{0.0041} & \textbf{0.0042}                                                    & \textbf{0.0476}                                          & N/A                                                           \\ \cline{2-6} 
                                                                              & \begin{tabular}[c]{@{}c@{}}MLFEM/ \\ FEM\end{tabular}      & 0.4954          & 0.0837                                                             & \textbf{0.0160}                                          & N/A                                                           \\ \hline
\multirow{3}{*}{\begin{tabular}[c]{@{}c@{}}Well \\ Classified\end{tabular}}   & \begin{tabular}[c]{@{}c@{}}Grad-CAM/ \\ MLFEM\end{tabular} & \textbf{0.0008} & \textbf{0.0068}                                                    & N/A                                                      & \textbf{0.0340}                                               \\ \cline{2-6} 
                                                                              & \begin{tabular}[c]{@{}c@{}}Grad-CAM/ \\ FEM\end{tabular}   & 0.9174          & 0.4664                                                             & N/A                                                      & 0.8683                                                        \\ \cline{2-6} 
                                                                              & \begin{tabular}[c]{@{}c@{}}MLFEM/ \\ FEM\end{tabular}      & \textbf{0.0009} & \textbf{0.0294}                                                    & N/A                                                      & \textbf{0.0255}                                               \\ \hline
\end{tabular}
 \caption{The p-value of the Mann-Whitney U test. If the Kruskal-Wallis H test suggested no significant difference between the explanation methods, the post-hoc Mann-Whitney U test was not conducted hence the ``N/A''}
    \label{tab:Mann-Whitney}
    \vspace{-0.5cm}
\end{table}

\begin{table}[ht!]
    \centering
    \footnotesize
    \begin{tabular}{|c|c|c|c|}
        \hline
        \rowcolor[HTML]{C0C0C0}
        \backslashbox{{\begin{tabular}[c]{@{}c@{}}Distortion\\ +\\ Classification\end{tabular}}}{\begin{tabular}[c]{@{}c@{}}Explanation \\ Methods\end{tabular}} & Grad-CAM & MLFEM & FEM \\\hline 
        Combined Distortion + Poor     & $\mathbf{2.0940\pm0.3685}$ & $1.9284\pm0.2635$ & $1.9611\pm0.2627$ \\ \hline
        Combined Distortion + Well     & $2.9190\pm0.9368$ & $\mathbf{3.2423\pm0.9088}$ & $2.9360\pm0.9171$ \\ \hline
        Additive Gaussian Noise + Poor & $\mathbf{2.1304\pm0.4061}$ & $1.9910\pm0.2665$ & $1.9108\pm0.2452$ \\ \hline
        Additive Gaussian Noise + Well & $2.7688\pm0.8902$ & $\mathbf{3.2471\pm0.8902}$ & $2.8870\pm0.8670$ \\ \hline
        Gaussian Blur + Poor           & $\mathbf{2.2714\pm0.3689}$ & $1.8681\pm0.2490$ & $2.0510\pm0.2490$ \\ \hline
        Gaussian Blur + Well           & $2.9848\pm0.9706$ & $3.1657\pm0.9367$ & $2.9521\pm0.9672$ \\ \hline
        Uni. Rand. Bright. Shift + Poor      & $1.9741\pm0.2775$ & $1.8973\pm0.2551$ & $1.9662\pm0.2834$ \\ \hline
        Uni. Rand. Bright. Shift + Well      & $2.9775\pm0.9151$ & $\mathbf{3.3508\pm0.8353}$ & $2.9629\pm0.8890$ \\ \hline
    \end{tabular}
    \caption{Mean MOS of explanation methods for each distortion type and classification category. The bold values indicate the best-performing explanation method based on the mean of MOS. Rows without bold values indicate no statistical significance.}
    \label{tab:SA-overall}
    \vspace{-0.8cm}
\end{table}

\subsubsection{Correlation Between MOS and Automatic Metrics.} \label{correlation}
The Spearman's rank correlation coefficient (SRCC) values between MOS and automatic metrics IAUC and DAUC can be seen in Tab. \ref{tab:cor-overall}. For IAUC and DAUC, 1\%  of the pixels were inserted or deleted at each deletion or insertion step. 
Row 1 contains SRCC on the whole set of maps (900) and we state that there exists a positive correlation between MOS and IAUC and DAUC. The row 2, 3, and 4 contain the SRCC values per method. The highest values are highlighted in bold. MLFEM's MOS shows the highest correlation with both metrics. However, the correlations are not very strong, especially for DAUC, which aligns with the findings in \cite{xudarme2023stabilitycorrectnessplausibilityvisual} that indicate no significant correlation between automatic and user-involved metrics. Users tend to trust more on metrics that incorporate their input than automatic metrics. This lack of consensus underscores that identifying the best evaluation metrics for explanation methods remains an open question. 

% - user will trust more on user studies than automatic. 

% The validity of using MOS as a metric to evaluate explanation methods has been pursued by checking the correlation between MOS and already-defined metrics 

% , both automatic and semi-automatic which use Gaze Fixation Density Maps (GFDMs). We have calculated the correlation between MOS and 4 metrics, Deletion Area Under the Curve (DAUC) \cite{petsiuk2018rise}, Insertion Area Under the Curve (IAUC) \cite{DBLP:journals/corr/abs-2201-13291}

% Looking at Tab.~\ref{tab:cor-overall}, MOS and the metrics DAUC, IAUC, PCC, and SIM for different explanation methods (All, Grad-CAM, MLFEM, and FEM) reveals distinct patterns. The "All" combination of methods shows moderate correlations with DAUC (0.3947) and IAUC (0.6369), but low correlations with PCC (0.1080) and SIM (0.0692). Grad-CAM follows a similar trend with moderate correlations to DAUC (0.3455) and IAUC (0.5770), and low correlations with PCC (0.1070) and SIM (0.0838). MLFEM stands out with the highest correlations to DAUC (0.4535) and IAUC (0.7033), suggesting its stronger alignment with these metrics, though it also exhibits low correlations with PCC (0.1061) and SIM (0.0755). FEM, while showing moderate correlations with DAUC (0.4009) and IAUC (0.6249), has the lowest correlation with PCC (0.0234) and a negative correlation with SIM (-0.0658), indicating a potential inverse relationship. Overall, MLFEM proves to be the most robust method in relation to DAUC and IAUC, while all methods demonstrate consistently low correlations with PCC and SIM.

\begin{table}[ht!]
\centering
\footnotesize
\begin{tabular}{|l|l|l|}
\hline
\rowcolor[HTML]{C0C0C0}
         & DAUC                  & IAUC                     \\ \hline
Overall  & 0.4406 (0.0)          & 0.6078 (0.0)             \\ \hline
Grad-CAM & 0.3765 (0.0)          & 0.5448 (0.0)             \\ \hline
MLFEM    & \textbf{0.4979 (0.0)} & \textbf{0.6843 (0.0)}    \\ \hline
FEM      & 0.4541 (0.0)          & 0.5836 (0.0)             \\ \hline
\end{tabular}
\caption{Spearman's rank correlation coefficient between explanation methods and DAUC and IAUC, with p-values in parentheses. The coefficients are rounded to four decimal places, and the p-values are rounded to three decimal places. The highest correlation is shown in bold.}
\label{tab:cor-overall}
\vspace{-1cm}
\end{table}

\section{Conclusion and Future Work}
\label{sec:conclusion}
In this work, we proposed the use of the Mean Opinion Score (MOS), a metric commonly used in image quality assessment, as a user-centric metric for evaluating the quality of explanation maps of XAI methods. We compared MOS with ``automatic metrics'' IAUC and DAUC. MOS involves collecting subjective assessments from users, capturing their opinions on the explanations' clarity, usefulness, and satisfaction. Additionally, MOS is quick to compute and does not require users to answer extensive or complicated questionnaires.

Based on the preliminary work \cite{Zhukov_Benois-Pineau_Giot_2023} and the work of Ford et al. \cite{ford2022explaining}, we developed a methodology and conducted a psycho-visual experiment with 31 subjects, including both online and offline participants. We evaluated three feature attribution methods: Grad-CAM \cite{DBLP:journals/corr/SelvarajuDVCPB16}, MLFEM \cite{bourroux:hal-03689004}, and FEM \cite{ahmedasiffuad:hal-02963298}. These methods were applied to intentionally distorted images (with additive Gaussian noise, Gaussian blur, and uniform random brightness shift) classified using ResNet50 \cite{he2015deep} on a subsample of the SALICON dataset \cite{jiang2015salicon}.

Our statistical analysis allowed us to filter out subjects with the most discrepant opinions. It then revealed that both offline and online participants exhibited similar behavior, allowing us to combine their responses for the MOS computation after filtering out the outliers. The comparison of MOS indicates the possibility of differentiating the performances of explanation methods. 
Therefore, we conclude that it can be used as the evaluation metric for XAI methods. 

Building on previous work \cite{bourroux:hal-03689004} which demonstrates the MLFEM's effectiveness compared to Grad-CAM and FEM, our findings further reveal MLFEM's superior ability to explain well-classified images in terms of MOS. Interestingly, Grad-CAM consistently performed well for poorly classified images.  

Additionally, Spearman's rank correlation coefficient showed that MOS measured for the MLFEM explainer had the highest correlation with automatic metrics IAUC and DAUC. Overall, the positive correlation between this user-centric metric and automatic metrics remains limited, as it was also shown in \cite{xudarme2023stabilitycorrectnessplausibilityvisual} for other user-centered semi-automatic metrics involving Gaze Fixation Density Maps. This lack of consensus highlights that finding the evaluation metrics that users will trust for explanation methods remains an open question that requires further investigation.

\bibliographystyle{splncs04}
\bibliography{references}

\end{document}